\documentclass[11pt,a4paper]{article}
\usepackage[hyperref]{emnlp2020}
\usepackage{times}
\usepackage{latexsym}
\usepackage{microtype}
\usepackage{xspace}
\usepackage{graphicx}
\usepackage{float}
\usepackage{printlen}
\usepackage[misc]{ifsym}
\usepackage{enumitem}
\usepackage{booktabs}

\aclfinalcopy 

\newcommand{\ours}{\texttt{JOELIN}\xspace}

\title{TEST\_POSITIVE at W-NUT 2020 Shared Task-3:\\ Joint Event Multi-task Learning for Slot Filling in Noisy Text}

\author{Chacha Chen \textsuperscript{1}\\Penn State \\chachachen@psu.edu \And Chieh-Yang Huang  \\Penn State \\ chiehyang@psu.edu \AND   Yaqi Hou \\UNC at Chapel Hill\\yaqi.hou@unc.edu \And Yang Shi \\NC State\\yshi26@ncsu.edu
   \And  Enyan Dai \\Penn State \\emd5759@psu.edu \And Jiaqi Wang\textsuperscript{*}  \\Penn State\\jqwang@psu.edu  }

\begin{document}
\maketitle
\begin{abstract}

The competition of extracting COVID-19 events from Twitter is to develop systems that can automatically extract related events from tweets. The built system should identify different pre-defined slots for each event, in order to answer important questions (e.g., \textit{Who is tested positive? What is the age of the person? Where is he/she?}). To tackle these challenges, we propose the \underline{Jo}int \underline{E}vent Mu\underline{l}ti-task Learn\underline{in}g (\ours) model. Through a unified global learning framework, we make use of all the training data across different events to learn and fine-tune the language model.  Moreover, we implement a type-aware post-processing procedure using named entity recognition (NER) to further filter the predictions. \ours outperforms the BERT baseline by $17.2\%$ in micro F1.\footnote{\url{https://github.com/Chacha-Chen/JOELIN}}


\end{abstract}

%
\section{Introduction}


In this work, we report the system architecture and results of the team TEST\_POSITIVE in the competition of W-NUT 2020 sharred Task-3: extracting COVID-19 event from Twitter. 

Since February 2020, the pandemic COVID-19 has been spreading all over the world, posing a significant threat to mankind in every aspect. The information sharing about a pandemic has been critical in stopping virus spreading. With the recent advance of social networks and machine learning, we are able to automatically detect potential events of COVID cases, and identify key information to prepare ahead.



We are interested in COVID-19 related event extraction from tweets. With the prevalence of coronavirus, Twitter has been a valuable source of news and information. Twitter users share COVID-19 related topics about personal narratives and news on social media \citep{muller2020covid}. The information could be helpful for doctors, epidemiologists, and policymakers in controlling the pandemic. However, manual extracting useful information from tremendous amount of tweets is impossible. Hence, we aim to develop a system to automatically extract structured knowledge from Twitter.

Extracting COVID-19 related events from Twitter is non-trivial due to the following challenges: \\
(1) \textbf{How to deal with limited annotations in heterogeneous events and subtasks?}. The creation of the annotated data relies completely on human labors, and thus only a limited amount of data can be obtained in each event categories. There are a variety types of events and subtasks.
Many existing works solve these low resource problem by different approaches, inlcuding crowdsourcing \citep{muller2020covid,finin2010annotating,potthast2018crowdsourcing}, unsupervised training \citep{xie2019unsupervised,hsu2017unsupervised}, or multi-task learning \citep{zhang2017survey,pentyala2019multi}. Here we adopt multi-task training paradigm to benefit from the inter-event and intra-event (subtasks) information sharing. In this way, \ours learns a shared embedding network globally from all events data. In this way, we implicitly augment the dataset by global training and fine-tuning the language model.

\begin{figure}[h]
\centering
\includegraphics[width=0.9\columnwidth]{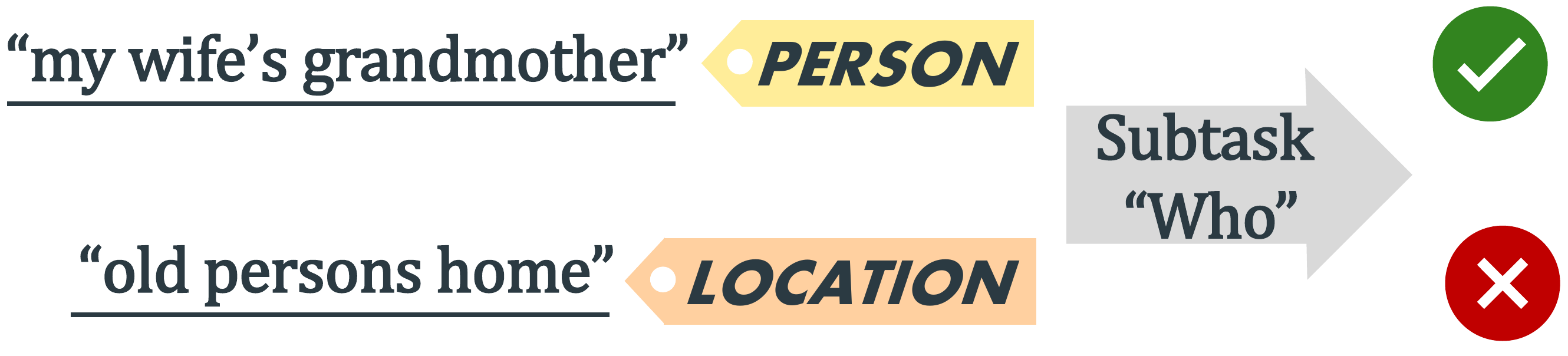}
\caption{Illustration of NER-based post-processing. }
\label{fig:ner-post-processing}
\end{figure}

\noindent(2) \textbf{How to make type-aware predictions?} Existing work \citep{zong2020extracting} did not encode the information of different subtask types into the model, while it could be useful in suggesting the candidate slot entity type. In order to make type-aware predictions, we propose a NER-based post-processing procedure in the end of \ours pipeline. We use NER to automatically tag the candidate slots and remove the candidate whose entity type does not match the corresponding subtask type. For example, as shown in Figure~\ref{fig:ner-post-processing}, in subtask ``Who'', ``my wife's grandmother'' is a valid candidate slot, while ``old persons home'', tagged as location entity, would be replaced with ``Not Specified'' during the post-processing. 

In summary, \ours is enabled by the following technical contributions:\\
  \indent  $\bullet$ \textbf{A joint event multi-task learning framework for different events and subtasks.} With the unified global training framework, we train and fine-tune the language model across all events and make predictions based on multi-task learning to learn from limited data. \\
 \indent  $\bullet$ \textbf{A NER-based type-aware post-processing approach.} We leverage NER tagging on the model predictions and filter out wrong predictions based on subtask types. In this way, \ours benefits from subtask type prior knowledge and further boosts the performance.

    






\section{Related Work}
\paragraph*{Event Extraction from Twitter} 

Impressive efforts have been made to detect events from Twitter. Existing works include domain specific event extraction and open domain event extraction. For domain specific extraction, approaches mainly focus on extracting a particular type of events, including natural disasters \citep{sakaki2010earthquake}, traffic events \citep{dabiri2019developing}, user mobility behaviors \citep{yuan2013and}, and etc. The open domain scenario is more challenging and usually relies on unsupervised approaches. Existing works usually create clusters with event-related keywords \citep{parikh2013events}, or named entities \citep{mcminn2015real,edouard2017graph}. Additionally, \citet{ritter2012open} and \citet{zhou2015unsupervised} design general pipelines to extract and categorize events in supervised and unsupervised manner respectively.



Different from previous works, we deal with COVID-19 related event extraction in particular. \citet{zong2020extracting} provide a BERT baseline for the same task. But we create a unified framework to learn simultaneously for different categories of events and subtasks.


\paragraph*{Type-aware Slot Filling}
\citet{yang2016cmuml} formulate entity type constraints and use integer linear programming to combine them with relation classification. \citet{adel2019type} propose to integrate entity and relation classes in convolutional neural networks and learn the correlation from data. 
We propose a NER-based post-processing technique for type-aware slot filling.
By filtering out entity mis-matched predictions, \ours can efficiently boost the performance with minimum hand-crafted rules.

\paragraph*{COVID-19 Twitter Analysis}
With the quarantine situation, people can share thoughts and make comments about COVID-19 on Twitter. It has become a research source for researchers to explore and study. \citet{singh2020first} show that Twitter conversations indicate a spatio-temporal relationship between information flow and new cases of COVID-19. There is some work about COVID-19 datasets. \citet{banda2020large} provide a large-scale curated dataset of over 152 million tweets. \citet{chen2020covid} collect tweets and forms a multilingual COVID-19 Twitter dataset. Based on the collected data, \citet{jahanbin2020using} propose a model to predict COVID-19 breakout by monitoring and tracking information on Twitter. Though there are some works about COVID-19 tweets analyisis \citep{muller2020covid, jimenez2020coronavirus, lopez2020understanding}, the work about automatically extracting structured knowledge of COVID-19 events from tweets is still limited.

\section{Method}
In this section, we introduce our approach \ours and its data pre-processing and post-processing steps in detail. First, we pre-process the noisy Twitter data following the data cleaning procedures in \citet{muller2020covid}. Second, we train \ours and fine-tune the pre-trained language model end-to-end. Specifically, we design the \ours classifier in a joint event multi-task learning framework. Moreover, we provide four options of embedding types and ensemble the outputs with the highest validation score. Finally, we further utilize NER techniques to post-process our results with minimum hand-crafted rules.


\begin{figure}[h]
\centering
\includegraphics[width=1\columnwidth]{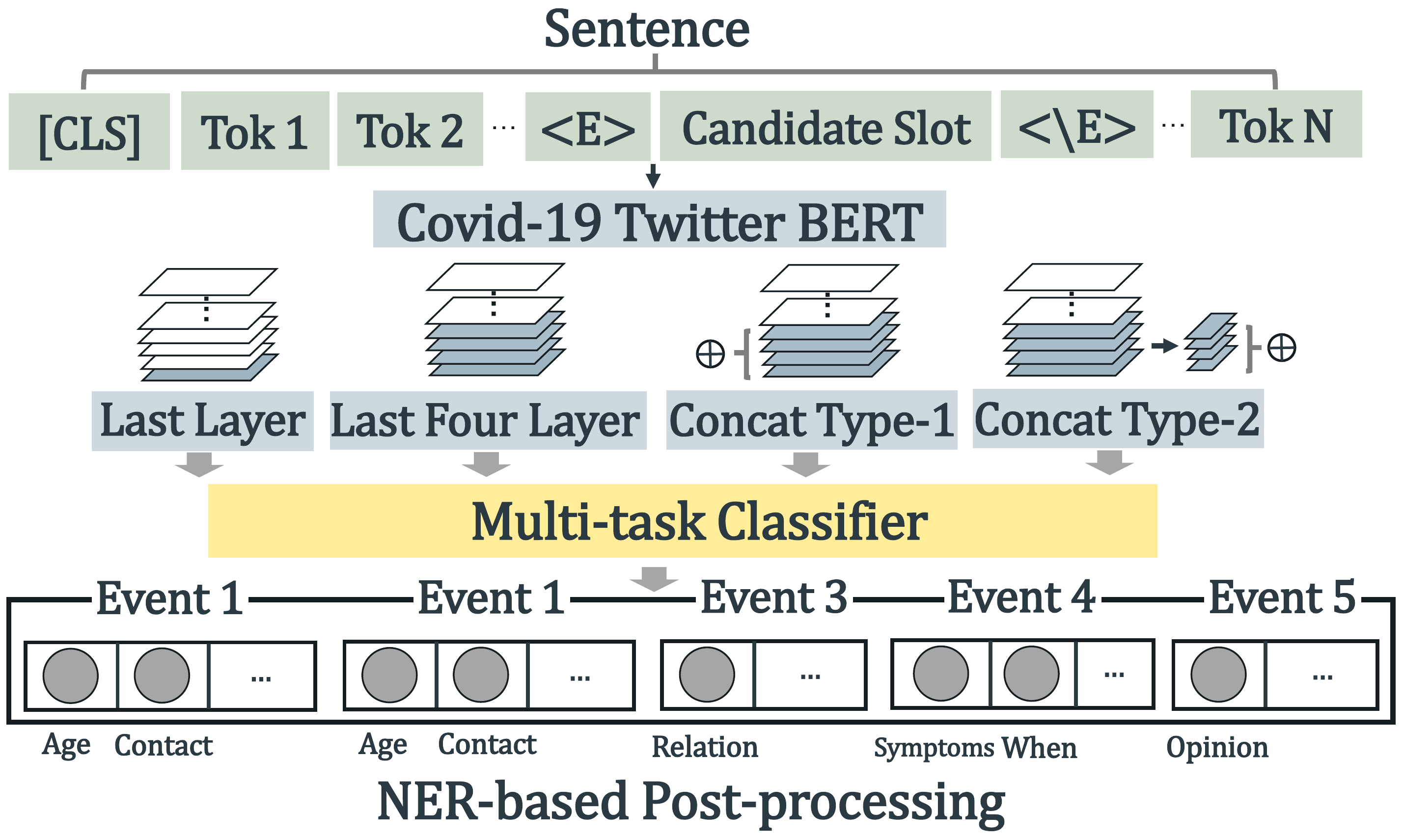}
\caption{Our approach comprises of 2 main components: (1) global language model across events and subtasks; (2) multi-task learning classifier. }
\label{model}
\label{fig:framework}
\end{figure}

\subsection{Data Pre-processing}
Prior to training, the original tweets are cleaned following \citet{muller2020covid}. The punctuations are standardized and unicode emoticons are expanded into textual ASCII representations\footnote{\url{https://pypi.org/project/emoji/}}.  All Twitter usernames are replaced with a special token \verb|<USER>| for pseudonymisation, URLs with \verb|<URL>|, and COVID-19 related tags, such as \#COVID19, \#coronavirus, \#COVID etc., with \verb|<COVID_TAG>|.
Note that the data cleaning step is designed as a hyper-parameter and can be on or off during the experiments.
 
We construct the training instance as follows. The annotated data is a collection of tweets. Each tweet is accompanied by hand-labeled candidate chunks. Each candidate chunk is extracted and sandwiched by a pair of tokens $<$E$>$ and $<$/E$>$.
The masked text, together with the annotated label, will then serve as one instance of the input.

\subsection{The \ours Model}
\ours consists of four modules as shown in Figure~\ref{fig:framework}: the pre-trained COVID Twitter BERT (CT-BERT) \citep{muller2020covid}, four different embedding layers, joint event multi-task learning framework with global parameter sharing, and the output ensemble module. 
\paragraph*{COVID Twitter BERT}
It has been a common practice that pre-trained language models, e.g., BERT~\citep{devlin2018bert} and RoBERTa~\citep{liu2019roberta}, are used for a supervised fine-tuning for specific downstream tasks. In this work, we use CT-BERT as \ours pre-trained language model. The CT-BERT is trained on a corpus of 160M tweets related to COVID-19. CT-BERT shows great improvement compared to BERT-LARGE and RoBERTa. We further fine-tune CT-BERT with the provided dataset.

\paragraph*{Feature Extraction} 
With the hidden representation of token $<$E$>$ given by CT-BERT, we further apply various choices of different feature extraction methods to choose the more useful features. Inspired by ~\citet{devlin2018bert}, we implemented the following four feature extraction methods:

    \noindent1. \underline{\textit{Last hidden layer}}: we directly use the last hidden layer of CT-BERT as our classifier input.\\
    2. \underline{\textit{Summation of last four}}: we sum the last four hidden layer outputs as the classifier input.\\
    3. \underline{\textit{Concatenation of last four (type-1)}}: we directly concatenate the last four layers, and flatten the vector before feeding it to the classifier.\\
    4. \underline{\textit{Concatenation of last four (type-2)}}: Each of last four layers is passed through a fully-connected layer and reduced to a quarter of its original hidden size. We flatten the vectors before passing through the classifier.
\paragraph*{Joint Event Multi-task Learning}
To tackle the challenge of limited annotated data, we apply a global parameter sharing model across all events. Specifically, we jointly learn and fine-tune the language embedding across different events and apply a multi-task classifier for prediction. As shown in Figure~\ref{fig:framework}, the language embedding as well as the feature extraction mechanism are jointly learned and fine-tuned globally. We then apply a fully-connected layer as our classifier for all the subtasks in different categories of events. In this way, \ours benefits from using data of all the events and their subtasks. Compared with training separate models for each event, joint training across different tasks significantly boosts the performance.

\paragraph{Model Ensemble} It has long been observed that ensembles of models boost overall performance. Hence, in this work, we train multiple models with different feature extraction approaches, and we select the top 5 models with best performance and ensemble them by majority voting.

\subsection{NER-based Post-processing}
We further filter our prediction based on NER for post-processing. Specifically, we use \textsf{spaCy}'s NER model\footnote{\url{https://spacy.io/}} to tag the predicted candidate slots. Then we compare the entity tag with the subtask. If the candidate tag does not match the subtask type, we invalidate the prediction by replacing it with \textit{``NOT SPECIFIED''}. For example, if the subtask is ``who'', we nullify those candidate slots whose tags are not related to persons, as shown in Figure~\ref{fig:ner-post-processing}.


\section{Experiments and Analysis}
\subsection{Dataset}
\vspace{-0.05in}
The dataset\footnote{\url{ https://github.com/viczong/extract_COVID19_events_from_Twitter}} is composed of annotated tweets sampled from January 15, 2020 to April 26, 2020. It contains 7,500 tweets for the following 5 events: (1) tested positive, (2) tested negative, (3) can not test, (4) death, and (5) cure and prevention. Each event contains several slot subtasks. 
\vspace{-0.05in}
\subsection{Implementation Details}
\label{sec:imple}
\vspace{-0.05in}
We randomly split the dataset into training and validation in a 80:20 ratio. 
The model is trained with the AdamW optimizer \cite{loshchilov2017decoupled} toward minimizing the binary cross entropy loss with batch size of 32 and learning rate of $2e$-$5$.
To deal with the class imbalance issue, we apply class weighting on the loss function. 
With grid-search, the best weight is 10 and 1 for positive and negative samples respectively.




\begin{table}[h!]
\centering \small
\begin{tabular}{lccc}
\toprule
Sub-task            & BERT  & CT-BERT & \ours  \\
\midrule
\multicolumn{4}{c}{TESTED POSITIVE }       \\
\midrule
age                 & 0.519 & 0.571   & 0.769 \\
close\_contact      & 0.262 & 0.333   & 0.420 \\
employer            & 0.394 & 0.391   & 0.453 \\
gender\_male        & 0.664 & 0.669   & 0.711 \\
gender\_female      & 0.635 & 0.698   & 0.779 \\
name                & 0.740 & 0.774   & 0.807 \\
recent\_travel      & 0.227 & 0.391   & 0.567 \\
relation            & 0.476 & 0.621   & 0.769 \\
when                & 0.571 & 0.571   & 0.741 \\
where               & 0.560 & 0.631   & 0.660 \\
\midrule
\multicolumn{4}{c}{TESTED NEGATIVE   }      \\
\midrule
age                 & 0.000 & 0.750   & 0.750 \\
close\_contact      & 0.000 & 0.133   & 0.133 \\
gender\_male        & 0.479 & 0.660   & 0.706 \\
gender\_female      & 0.214 & 0.649   & 0.766 \\
how\_long           & 0.000 & 0.400   & 0.800 \\
name                & 0.519 & 0.646   & 0.675 \\
relation            & 0.449 & 0.720   & 0.784 \\
when                & 0.000 & 0.471   & 0.471 \\
where               & 0.372 & 0.578   & 0.651 \\
\midrule
\multicolumn{4}{c}{CAN NOT TEST      }      \\
\midrule
relation            & 0.516 & 0.608   & 0.771 \\
symptoms            & 0.517 & 0.704   & 0.757 \\
name                & 0.382 & 0.545   & 0.550 \\
when                & 0.000 & 0.000   & 0.000 \\
where               & 0.509 & 0.500   & 0.638 \\
\midrule
\multicolumn{4}{c}{DEATH             }      \\
\midrule
age                 & 0.727 & 0.722   & 0.789 \\
name                & 0.642 & 0.715   & 0.774 \\
relation            & 0.378 & 0.646   & 0.680 \\
symptoms            & 0.000 & 0.000   & 0.444 \\
when                & 0.633 & 0.605   & 0.690 \\
where               & 0.483 & 0.613   & 0.628 \\
\midrule
\multicolumn{4}{c}{CURE AND PREVENTION }      \\
\midrule
opinion             & 0.520 & 0.573   & 0.627 \\
what\_cure          & 0.583 & 0.671   & 0.671 \\
who\_cure           & 0.389 & 0.515   & 0.545\\
\midrule
\midrule
micro avg. F1    &0.576 &0.647 &\textbf{0.696}\\
\bottomrule
\end{tabular}
\vspace{-0.05in}
\caption{Overall performance of \ours compared with BERT and CT-BERT on validation data. The results are reported with F1 score.}
\label{tab:overall}
\end{table}
\vspace{-0.05in}
\subsection{Results and Discussion}
\vspace{-0.05in}
We evaluate \ours with BERT and CT-BERT baselines. We measure the performance of different models with F1 score and micro F1 score, in consideration of imbalanced sample sizes. The overall results are shown in Table~\ref{tab:overall}. Compared with the performance of BERT~\citep{zong2020extracting} and CT-BERT~\citep{muller2020covid}, \ours significantly outperforms the best baseline CT-BERT by $7.6\%$ in micro F1. In terms of performance on subtasks, \ours outperforms the best baseline CT-BERT by up to $44.9\%$ in recent travel of event TESTED POSITIVE. The performance gains of \ours are attributed to the well-designed joint event multi-task learning framework and the type-aware NER-based post-processing.

\begin{table}[h!]
\centering \small
\begin{tabular}{lc}
\toprule
   Model       &       Micro F1      \\ 
    \midrule
\ours-P                 &   0.488           \\
\ours                   &   \textbf{0.511}           \\
\bottomrule
\end{tabular}
\vspace{-0.05in}
\caption{Ablation model comparison on test data.} 
\vspace{-0.1in}
\label{tab:ablation}
\end{table}

\vspace{-0.05in}
\subsection{Ablation Study}
\vspace{-0.05in}
We conduct an ablation study to understand the contribution of type-aware post-processing in \ours. We remove the post-processing step as a reduced model (\ours-P) and compare the micro F1 scores. 
As shown in Table~\ref{tab:ablation}, \ours has better micro F1 score in comparison with the reduced model \ours-P. It supports the claim that our proposed type-aware post processing with NER can significantly boost the performance.

\section{Conclusion}
In this work, we build \ours upon a joint event multi-task learning framework. We use NER-based post-processing to generate type-aware predictions. The results show \ours significantly boosts the performance of extracting COVID-19 events from noisy tweets over BERT and CT-BERT baselines. In the future, we would like to extend \ours to open domain event extraction tasks, which is more challenging and requires a more general pipeline. 

\clearpage
\bibliographystyle{acl_natbib}
\bibliography{emnlp2020}



\end{document}